\documentclass[12pt]{iopart}
\expandafter\let\csname equation*\endcsname\relax
\expandafter\let\csname endequation*\endcsname\relax
\usepackage{cite}
\usepackage{amsmath}
\usepackage{amssymb}
\usepackage{amsfonts}
\usepackage{mathtools}
\usepackage{graphicx}
\usepackage{textcomp}
\usepackage{xcolor}
\usepackage{citesort}
\usepackage[]{algorithm2e}
\newcommand{\ignore}[1]{}
\usepackage{algorithmic}
\usepackage[utf8]{inputenc}
\usepackage{stackengine}
\usepackage{bm}

\begin{document}

\title{A Neural Network for Determination of Latent Dimensionality in Nonnegative Matrix Factorization}

\author{Benjamin T. Nebgen, Raviteja Vangara, Miguel A. Hombrados-Herrera, Svetlana Kuksova, Boian S. Alexandrov}

\address{Los Alamos National Laboratory, Los Alamos, NM, 87544, United States}
\ead{bnebgen@lanl.gov}
\ead{boian@lanl.gov}
\vspace{10pt}
\begin{indented}
\item[]April 2020
\end{indented}

\begin{abstract}
Non-negative Matrix Factorization (NMF) has proven to be a powerful unsupervised learning method for uncovering hidden features in complex and noisy data sets with applications in data mining, text recognition, dimension reduction, face recognition, anomaly detection, blind source separation, and many other fields. An important input for NMF is the latent dimensionality of the data, that is, the number of hidden features, $K$, present in the explored data set. Unfortunately, this quantity is rarely known \emph{a priori}. The existing methods for determining latent dimensionality, such as Automatic Relevance Determination (ARD), are mostly heuristic and utilize different characteristics to estimate the number of hidden features. However, all of them require human presence to make a final determination of $K$. Here we utilize a supervised machine learning approach in combination with a recent method for model determination, called NMFk, to determine the number of hidden features automatically. NMFk performs a set of NMF simulations on an ensemble of matrices, obtained by bootstrapping the initial data set, and determines which $K$ produces stable groups of latent features that reconstruct the initial data set well. We then train a Multi-Layer Perceptron (MLP) classifier network to determine the correct number of latent features utilizing the statistics and characteristics of the NMF solutions, obtained from NMFk. In order to train the MLP classifier, a training set of 58,660 matrices with predetermined latent features were factorized with NMFk. The MLP classifier in conjunction with NMFk maintains a greater than 95\% success rate when applied to a held out test set. Additionally, when applied to two well-known benchmark data sets, the swimmer and MIT face data, NMFk/MLP correctly recovered the established number of hidden features. Finally, we compared the accuracy of our method to  the ARD, AIC and Stability-based methods.
\end{abstract}

\section{Introduction}
In today's world, great volumes of data are generated from sources such as social networks, computer macro-simulations, sensor arrays, engineering activities, communication media, sequencing of human (or other) genomes, and others \cite{dong2013big,cuzzocrea2011analytics}. data sets contain values of directly observable values while the processes that cause the manifestation of those observations remain hidden or latent \cite{everett2013introduction,franke2016statistical}. Usually, the latent variables (or features) producing the data are either impossible to measure directly or are simply unknown. Intelligently utilizing the data, for example in data-driven science, decision making, or emergency response, requires understanding the primary processes that are the cause of the studied phenomenon and therefore requires identification of the latent variables which dictate the observations. This mandates the ability to extract from data understandable latent variables. Additionally, this process has the added benefit of reducing the dimension of the data set since each observable can be expressed as the combination of a smaller number of latent variables. This type of dimension reduction is frequently the subject of unsupervised learning \cite{barlow1989unsupervised} and utilizes many methods such as factor analysis \cite{spearman1961general}, subspace clustering \cite{parsons2004subspace}, Principle Component Analysis (PCA)\cite{jolliffe2016principal}, Independent Component Analysis (ICA) \cite{amari1996new}, and Non-negative Matrix Factorization (NMF) \cite{paatero1994positive,Nasrin_ravi}.

Importantly, the non-negativity constraint in NMF guarantees that the extracted latent variables will be physically interpretable \cite{cichocki2009nonnegative} because NMF learns parts-based representations of data \cite{lee1999learning}. Indeed, when only addition but not subtractions are permissible, reproducing a data set requires the extracted latent variables to be parts of the original data, thus making them easy to understand and interpret. Many state variables, e.g., density, energy, spectra, population, etc., are naturally non-negative and many others can be examined as non-negative via suitable transformations.
\begin{figure}[htbp]
\centering
\includegraphics[width=0.7\textwidth]{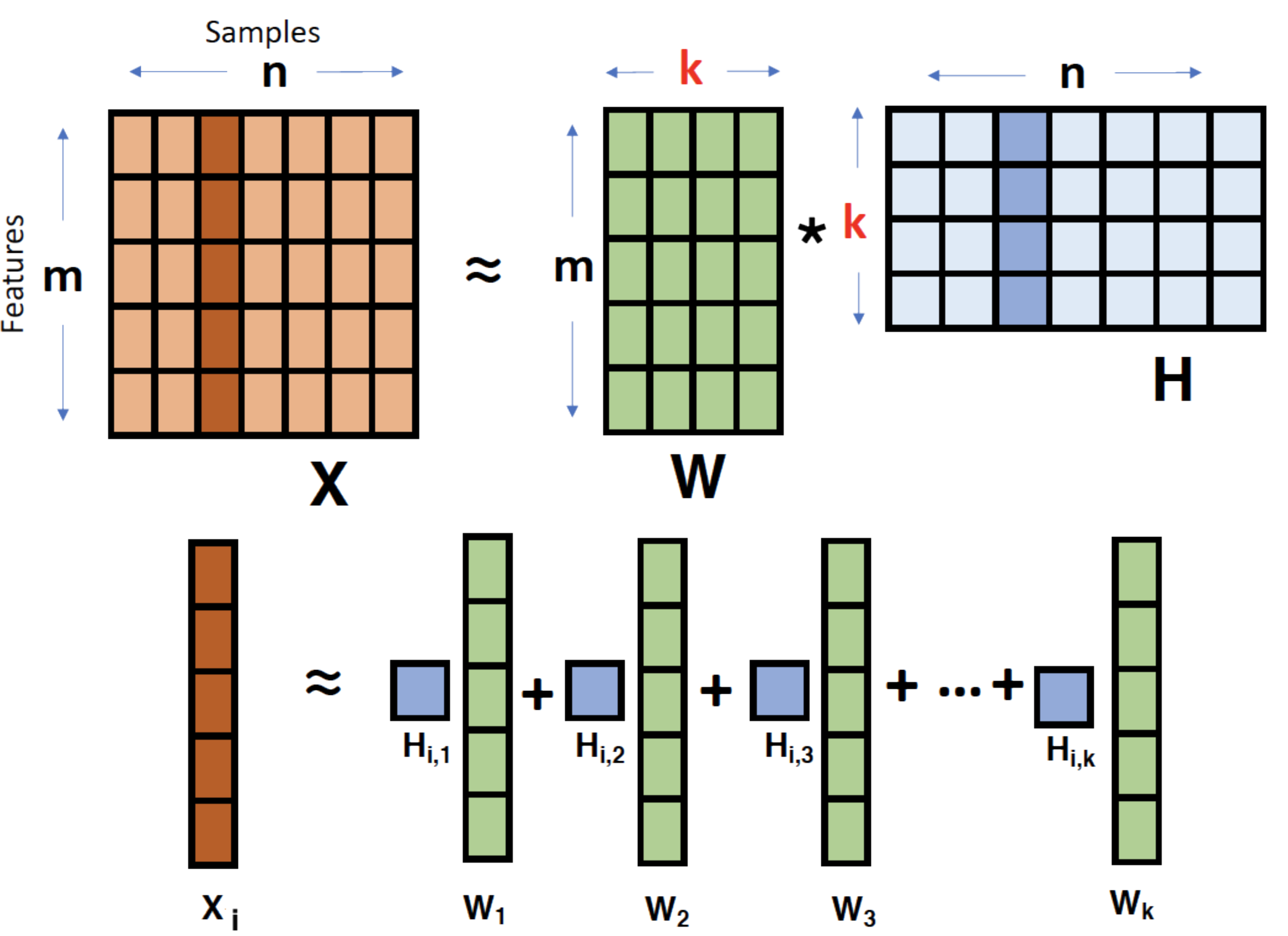}
\caption{An illustration of the non-negative matrix factorization of two low-rank matrices $\bf{W}$ and $\bf{H}$ with inner dimension $K$. Each column of the data $\bf{X}$ (a sample) is represented as a linear combination of the basis vectors: the columns of $\bf{W}$ with their their corresponding weights in $\bf{H}$.}
\label{fig:NMF}
\end{figure}

The usual interpretation of NMF is as a method for low-rank matrix approximation of the observed data-matrix $\bf{X}$, with size $m\times n$, by two unknown non-negative matrices, $\bf{W}$, with size $m\times K$, and $\bf{H}$, with size $K\times n$: $\bf{X}\approx \bf{WH}$. Both $\bf{W}$ and $\bf{H}$ contain one small dimension, $K$, which is the number of latent variables, as shown in Figure \ref{fig:NMF}. This approximation is performed through a non-convex minimization with a given distance metric, $||...||_{dist}$ :
\begin{equation}\label{distMet}
min||\bf{X}_{ij}-\sum^{K} _{s=1}\bf{W}_{is}\bf{H}_{sj}||_{dist}
\end{equation}
This minimization is constrained by the non-negativity of $\bf{W}$ and $\bf{H}$: $\bf{W}_{is}\geq0$; $\bf{H}_{sj}\geq0$. NMF has proven very useful for face recognition, text recognition\cite{kysenko2012gpu}, dimension reduction \cite{liu2008reducing}, unsupervised learning\cite{bertrand2008unsupervised}, anomaly detection\cite{abdel2016nmf,allan2008anomaly}, Blind Source Separation\cite{battenberg2009accelerating,vangaradiffusion}, and other problems. \cite{cichocki2009nonnegative}. Importantly, NMF is underpinned by a well-defined statistical generative model of superimposed components that, when the distance metric $||...||_{dist}$ is the Euclidean distance, can be treated as a Gaussian mixture model \cite{fevotte2009nonnegative}. In this case, the NMF algorithm is equivalent to the expectation-maximization (EM) algorithm \cite{dempster1977maximum} developed to find the maximum likelihood estimates of parameters in statistical models, where the model depends on latent variables \cite{tan2012automatic}. In this probabilistic interpretation of NMF, the observables  $x_1, x_2, ..., x_n$ ($x_i$ is a column vector of $\bf{X}$ with $m$ elements), are generated by $K$ latent variables, $h_1, h_2,..., h_{K}$. Specifically, each observable $x_i$ is generated from a probability distribution with mean  $\langle x_i \rangle=\sum^{K} _{ s=1}\bf{W}_{is}h_{s}$, where $K$ is the number of latent variables. The influence of $h_s$ on $x_i$ is through the basis patterns of the considered phenomenon $w_{:s}$ represented by the columns of $\bf{W}$ \cite{lee1999learning}. 
 
The NMF minimization requires the latent dimensionality $K$, to be known \emph{a priori}. Indeed, when $K$ is already known, the NMF algorithm itself is sufficient to conduct the constrained minimization and to extract the desired latent variables. However, the latent dimensionality is usually \emph{a priori} unknown even though the value of $K$ is crucial for the most of the NMF's applications. If $K$ is chosen to be too small the fit to the data-matrix $X$, will be poor. In the opposite situation where $K$ is chosen to be too large, NMF will over fit the data, resulting in incorrect latent variables.

\begin{figure*}[htbp]
\centering
\includegraphics[width=0.8\textwidth]{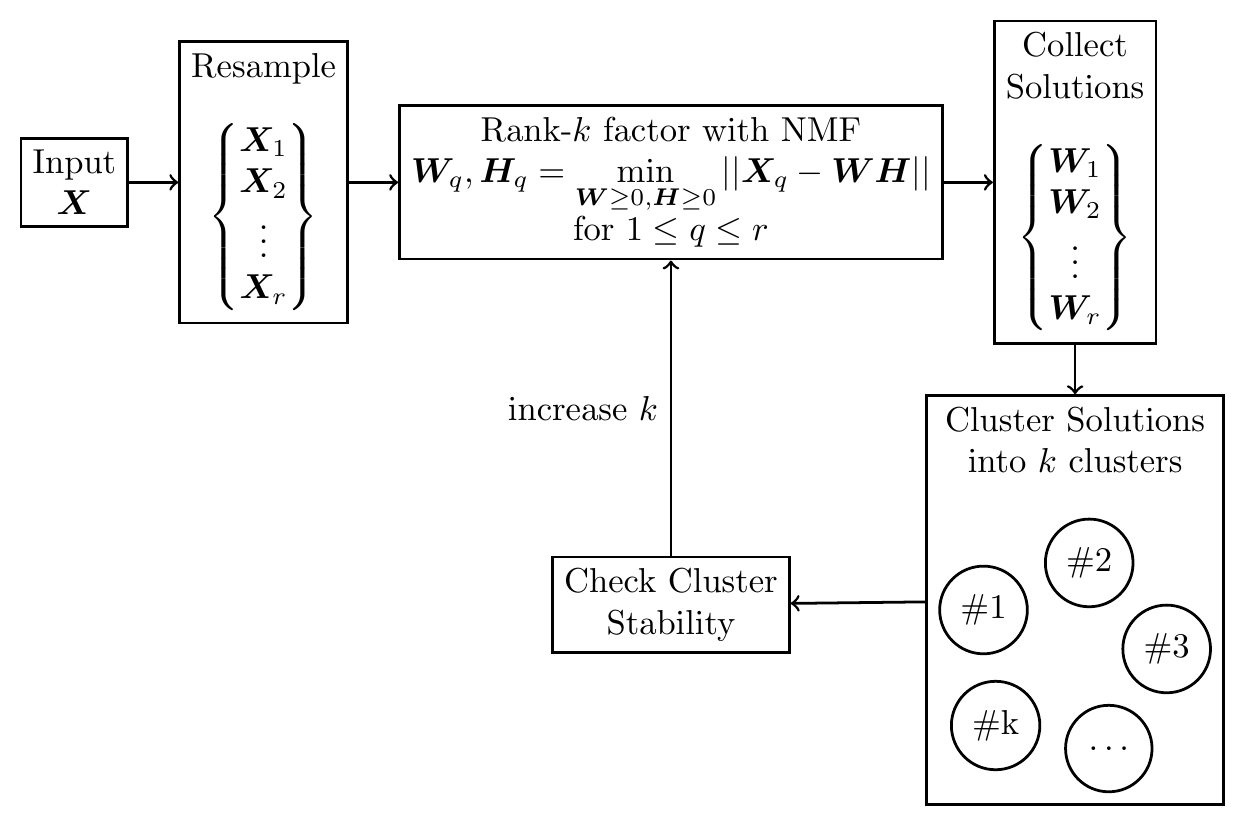}
\caption{Schematic of the NMFk procedure. The initial data set $\bf{X}$ is re-sampled into an ensemble of matrices $\left\{\bf{X}_i\right\}$. Traditional NMF is applied to each of these matrices and the resulting columns of $\bf{W}$ are clustered using K-means clustering.} 
\label{fig:NMFk}
\end{figure*}

All existing methods for determination of the latent dimensionality $K$ are, in general, heuristics. Brunet Et al. \cite{Brunet4164} proposed a method based on stability of NMF solution, utilized to identify the number of clusters in the observational matrix, $X$. The most well-known method is based on the automatic relevance determination (ARD) introduced by MacKay in \cite{mackay1994automatic}, and applied later for PCA by Bishop\cite{bishop1999bayesian}. An ARD protocol for NMF has been introduced by Fevotte and Tan \cite{fevotte2009nonnegative} and by Morup and Kai \cite{morup2009tuning}. A recent model determination technique, called NMFk, \cite{alexandrov2013deciphering,alexandrov2014blind,chennupatidistributed} has been successfully used to decompose the biggest collection of human cancer genomes \cite{alexandrov2013signatures}. NMFk complements classical NMF with custom clustering and Silhouette statistics \cite{rousseeuw1987silhouettes}. NMFk finds a trade-off between the stability of extracted latent variables from several NMF-minimizations at a given $K$, and the accuracy of the reconstructed data. NMFk uses re-sampling to create ensemble of initial matrices, $\bf{X} \rightarrow{(X_1, X_2, ..., X_r)}$, with a mean equal to the initial observational matrix, $X$. Additionally, the custom clustering algorithm searches for groups columns of $\bf{W}$, stable across the re-sampling of $\bf{X}$ for each explored $K$, as shown in Figure \ref{fig:NMFk}. 

In this paper we utilize a supervised learning technique for automating the determination of the correct number of latent variables. While there are other methods for determining $K$, as discussed above, these methods require human interpretation for a final decision. Here, we construct a Multi-Layer Perceptron (MLP) network to predict a $K$ value from the statistics produced by NMFk. By training the Multi-Layer Perceptron with 58,660 synthetically generated matrices with known $K$, this classifier network was able to predict the correct $K$ on a held out test set with 95\% accuracy. Additionally, this method shows success on a variety of data sets available from the literature. The layout of this paper is as follows: In Section \ref{sec:meth} we discuss NMFk, the MLP model, and details on training and inference. In Section \ref{sec:test} we apply the model to a variety of data sets including further synthetic data (Sections \ref{subsec:syn} and \ref{subsec:other}), the MIT face data set (Section \ref{subsec:MIT}), the swimmer data set (Section \ref{subsec:swim}), and the MNIST dataset of hand written Arabic numerals (Section \ref{subsec:mnist}). Finally, we conclude and examine future prospects for this method in Section \ref{sec:conc}.

\section{Methodology}
\label{sec:meth}

\subsection{NMFk for Model Determination}

Non-negative matrix factorization decomposes a given data matrix $\bf{X}$ into its latent variables. The data in $\bf{X}$ is expressed expressed as a linear combination of the columns of $\bf{W}$, with corresponding mixing weights in $\bf{H}$. Decomposition of the data matrix relies on the prior knowledge of number of latent features, $K$. Alexandrov et al \cite{alexandrov2013deciphering,alexandrov2014blind,chennupatidistributed} demonstrated a technique which identifies the number of hidden features by applying NMF to several instances of the re-sampled data $\left\{X_i\right\}$ and computing the stability of clusters constructed from the union of the columns of $\bf{W}$. The schematic for NMFk is shown in Figure \ref{fig:NMFk}. The data is re-sampled, i.e. perturbed with a noise from a uniform distribution, such that a distribution of input data is generated. Each re-sampling of $\bf{X}$ is factorized with NMF to produce a distribution of $\bf{W}$ factors for each value of $K$. The $K$ columns of $\bf{W}$ obtained from $\left\{X_i\right\}$, are grouped into $K$ clusters, as detailed in Figure \ref{fig:NMF}. The clustering in this case is K-means clustering, with fixed number of samples per cluster, and cosine distance as a similarity metric. The stability of these clusters is computed with the silhouette statistic \cite{rousseeuw1987silhouettes}, which for the ith point is given by: 
\begin{equation}\label{silh}
Silhouette (i) = \frac{b(i)-a(i)}{max\{a(i),b(i)\}} 
\end{equation}
Here, $a(i)$ is the average distance of point $i$ with that of the other points in the same cluster, and $b(i)$ is the minimum average distance from the ith point to points in a different cluster, minimized over clusters. A cluster well separated from other clusters will have a silhouette near 1, while a poorly separated cluster will have a silhouette near -1. Additionally, for each $K$ the Akaike Information Criterion (AIC)\cite{sakamoto1986akaike}, a widely used measure of statistical model quality, is computed using the following equation: 
\begin{equation}\label{AICEQ}
AIC = 2K + N_{t}\times ln \left( \frac{O}{N_{t}} \right)
\end{equation}
Where $N_t$ is the number of elements in $\bf{X}$ and $O$ is defined as the relative error in reconstruction:
\begin{equation}\label{OEQ}
O = \frac{\left\| \bf{X} - \bf{WH}\right\|}{\left\| \bf{X} \right\|}
\end{equation}

\begin{figure}[htbp]
\centering
\includegraphics[width=6in]{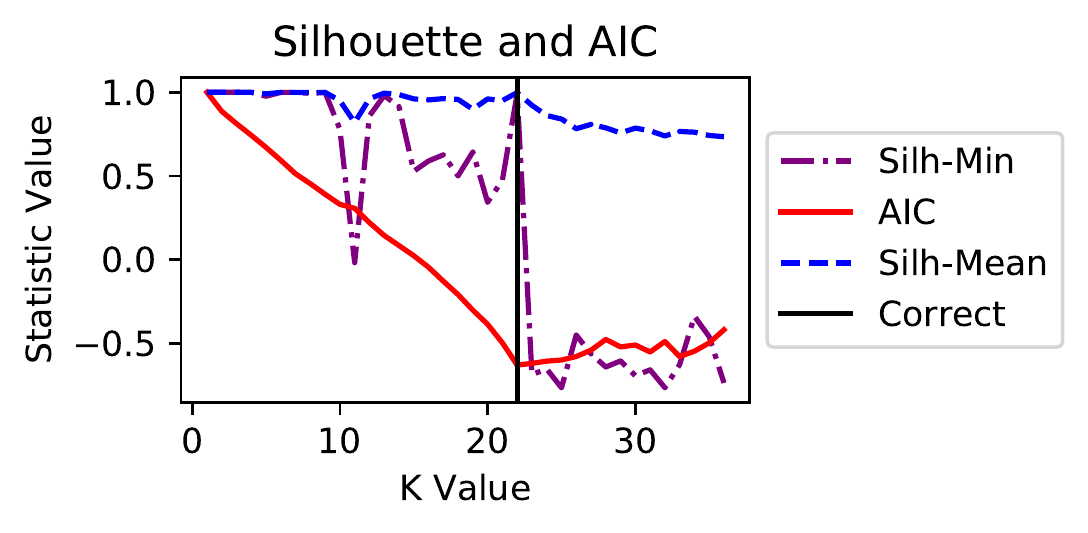}
\caption{AIC and Silhouette values computed for an example $X$ matrix using NMFk for a range of $K$ values. The black vertical line denotes the correct $K$ value, which is known since the data is synthetic. The left-hand axis applies to all three statistics.}
\label{SAexample}
\end{figure}

 When the NMFk algorithm is applied with a $K$ lower than the true number of features, the clusters of $\bf{W}$ columns are well separated resulting in minimum and average silhouette values near one. However, due to the poor reconstruction of the data the AIC value will be relatively large. These conditions can be observed on the left side of Figure \ref{SAexample}.When the $K$ exceeds the number of actual latent features in the data, NMFk produces a low AIC value, but the stability of $\bf{W}$ clusters drops. This is observed on the right side of Figure \ref{SAexample}. The correct value of $K$ is indicated by high minimum and average silhouette values as well as a (near) minimum AIC value, as illustrated by the black vertical line in Figure \ref{SAexample}. However, neither the highest $K$ with a minimum silhouette of 1 nor the $K$ with a minimum AIC value are reliable indicators of the correct $K$ value.

\subsection{Classifier Network} \label{secClass}

In an effort to fuse AIC and silhouettes to produce a more reliable and automated algorithm for selecting the correct number of latent variables, a Multi-Layer Perception (MLP) network was constructed utilizing the Sci-kit Learn toolbox \cite{scikit-learn}. An MLP network was chosen for this application since the model remains a constant size independent of training data set, unlike some models such as Support Vector Machines \cite{SVM1,SVM2,SVM3}. This allows for the generation of very large synthetic data sets for training. A hyper-parameter search over activation function, number of layers, optimizer, and tolerance parameters was utilized to determine the optimal network. To perform this search, 10\% of the initial training data was held out as a test set for hyper-parameter optimization, while 90\% was utilized to train the networks. This search resulted in a three layer network with 300, 200, and 100 neurons in each layer utilizing a rectified linear, or "re-lu" activation. An L2 regularization term of .01 was used to prevent over fitting. The network was trained to optimize a logistic loss function, as shown in Equ. \ref{lossEQ}, where $Y$ are the known class labels and $P$ are the predicted probabilities. 
\begin{equation}\label{lossEQ}
\begin{split}
L\left(\bf{Y},\bf{P}\right)=\frac{-1}{N}\sum_{i=1}^{N}\sum_{j=1}^{K}\bf{Y}_{i,k}log\left(\bf{P}_{i,k})\right) \\
\bf{Y}_{i,k}=
\begin{cases}
1 & $if sample i belongs to class k$ \\
0 & $otherwise$
\end{cases}
\end{split}
\end{equation}
During training, a tolerance factor of $2e10^{-4}$ was utilized to control both the adaptive learning rate and termination. The learning rate was reduced by a factor of five when two successive epochs failed to improve the accuracy of the network by more than the tolerance. Training was terminated when 10 successive epochs failed to increase the models accuracy by more than the tolerance factor. For training, the ADAM algorithm was utilized \cite{adam}. In training, a batch size of 200 data points was used, which struck a good balance between computational performance and network convergence. A model trained with 90\% of the training set was utilized for the remained of this paper, while models trained on 80\% splits of the data are shown in Section 2 of the Supplementary Information to confirm that this model performs in a average way. 

\begin{figure*}[htbp]
\centering
\includegraphics[width=6in]{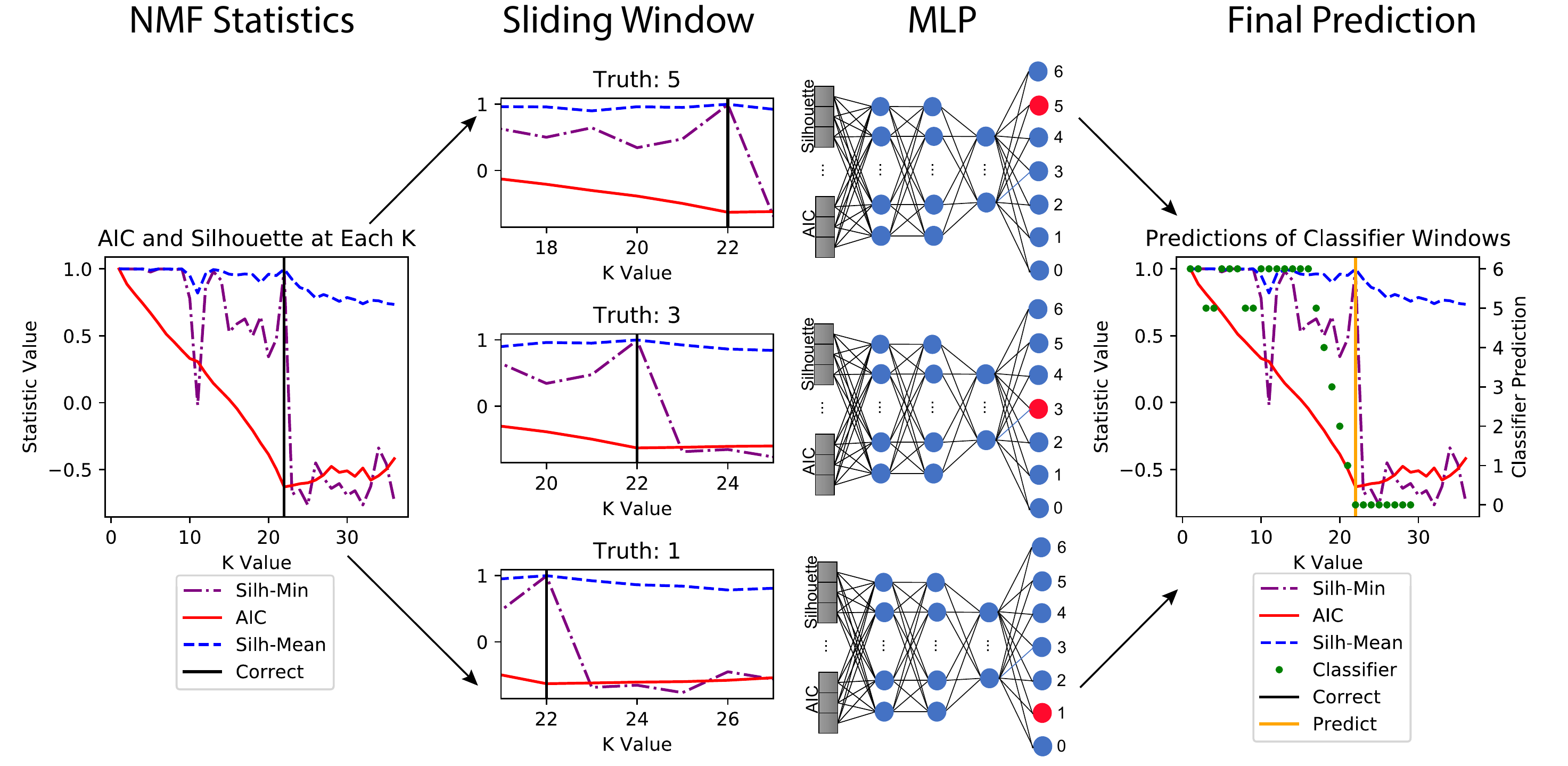}
\caption{Diagram NMFk paired with a MLP network. A sequence of AIC, minimum silhouette, and average silhouette as a function of $K$ is broken down into windows of 7 sequential $K$ values. These are labeled according to the relationship between the origin of the window and the true $K$ value. The network is then trained to correctly classify these different windows in 7 different classes. The value of AIC, minimum silhouette, and average silhouette are indicated by the left hand axis, while the classification of the MLP network (green dot) is indicated by the right hand axis.}
\label{ExplFig}
\end{figure*}

The training data for the MLP network is detailed in Figure \ref{ExplFig}. 58,660 synthetic matrices with a known latent dimensionality were generated using the method outlined in Section \ref{secTrain}. NMFk was then applied to each of these matrices with a variable number of features ranging from $K=1$ to $K=36$. At each $K$ value the AIC, minimum cluster silhouette, and maximum cluster silhouette were computed from the NMFk solutions. An example of these values can be seen on the left hand side of Figure \ref{ExplFig}. This sequence of AIC, minimum silhouette, and average silhouette as a function of $K$ was sub-sectioned into widows spanning 7 $K$ values. Each of these windows was given a label based upon the relationship between the origin of the window and the true number of latent features known from the data generation algorithm, as seen in Equ. \ref{labelEQ}. 

\begin{equation}\label{labelEQ}
Label = 
\begin{cases}
0 & K_{true} \leq K_{origin}\\
6 & K_{true} - K_{origin} \geq 6 \\
K_{true} - K_{origin} & otherwise
\end{cases}
\end{equation}

In this way, class 0 indicates that $K_{true}$ is either at or to the left of the origin, class 6 indicates that $K_{true}$ is at or to the right of the last point in the window, and classes 1-5 indicate that $K_{true}$ is that many points to the right of the sliding window. In this way, the MLP network can be passed over a large range of possible $K$ values with the MLP network indicating the correct number of features when it resides within the window.

\subsection{Training Data set} \label{secTrain}

In order to train the MLP model, a labeled data set of $\bf{X}$ matrices with a known number of features in each matrix needed to be constructed. This was done by building the underlying $\bf{W}$ and $\bf{H}$ matrices and multiplying them to form $\bf{X}$. Since the number of latent variables in $\bf{W}$ and $\bf{H}$ are known, the true $K$ can be used to label each matrix in this training data set. The algorithm sketched in Algorithm \ref{randAlg} was used to generate random matrices with known numbers of latent variables. The columns of $\bf{W}$ are Gaussian functions with randomized centers and widths. The first column of $\bf{W}$ is composed with no additional constraints. The second column of $\bf{W}$ is a random Gaussian with an enforced limit on the Pearson correlation with the initial Gaussian. For the generation of the data set used in this paper, three distinct correlations ranges between $w_1$ and $w_2$ were used: [0.2,0.4], [0.4,0.6], [0.6,0.8]. These limits were fed in as min-correlation and max-correlation in Algorithm \ref{randAlg} and ensure the existence of features with a variety of correlations levels. 

\begin{algorithm}
\SetAlgoLined
 \hspace*{\algorithmicindent} \textbf{Input} n, m, K, noise, min-correlation, max-correlation\\
 \hspace*{\algorithmicindent} \textbf{Output} matrix \bf{X} with size n$\times$m and K latent features \\
%\KwResult{$X=W\dot H\times e$}
 $\bf{W}(x,1)=e^{\frac{\left(x-a\right)^2}{b}}$; random a$\in \left[1,n\right]$,b$\in \left[1,\frac{n}{10}\right]$ \par
 \While{not min-correlation $< Pearson(\bf{W}(1),\bf{W}(2)) <$ max-correlation}{
  $\bf{W}(x,2)=e^{\frac{\left(x-a\right)^2}{b}}$; random a$\in \left[1,n\right]$,b$\in \left[1,\frac{n}{10}\right]$ \par
 }
 \For{i=3;K}{
  \While{not $(Pearson(\bf{W}(i),\bf{W}(j)) < 0.3 )_{\forall j < i}$}{
   $\bf{W}(x,i)==e^{\frac{\left(x-a\right)^2}{b}}$; random a$\in \left[1,n\right]$,b$\in \left[1,\frac{n}{10}\right]$}
  }
\bf{H}=exp(rand(K,m)\par
\bf{err}=random matrix n$\times$m in range [1-noise,1+noise]\par
$\bf{X}=\bf{WH} \times \bf{err}$
\caption{Algorithm for the construction of synthetic data matrices with known numbers of hidden features.}
\label{randAlg}
\end{algorithm}

Next, the remaining $K-2$ features were generated for each $W$ matrix. These features are similar random Gaussians with a maximum Pearson correlation of 0.3 to all other previously generated features. This correlation limit prevented the construction of $\bf{X}$ matrices with features that could not be resolved by NMF. Once all of the features of $\bf{W}$ were generated, $H$ was generated simply from an exponential applied to a uniform random distribution in the range [0.0,1.0]. Finally, a noise term was generated to introduce errors in the reconstruction of $\bf{W}$. Without the noise term, determining the  number of features would be trivial as it would be the lowest number of features with perfect reconstruction of $\bf{X}$. Just as with the correlation in $\bf{W}$ rows, three different noise levels were selected: 5\%, 10\%, and 20\%, which were fed into Algorithm \ref{randAlg} as the noise parameter. In practice, a random matrix of the same dimensionality of $\bf{X}$ was constructed with values in the range $[1-Noise,1+Noise]$, where $Noise$ is one of the previously defined values. Finally, $\bf{X}$ is assembled by taking the matrix product of the previously generate $\bf{W}$ and $\bf{H}$ with the element wise product of the noise matrix. All $\bf{W}$, $\bf{H}$, and $\bf{X}$ matrices were stored for later computation of silhouette and AIC. In total, 58,660 matrices were generated for training. 

\begin{figure}[htbp]
\centering
\includegraphics{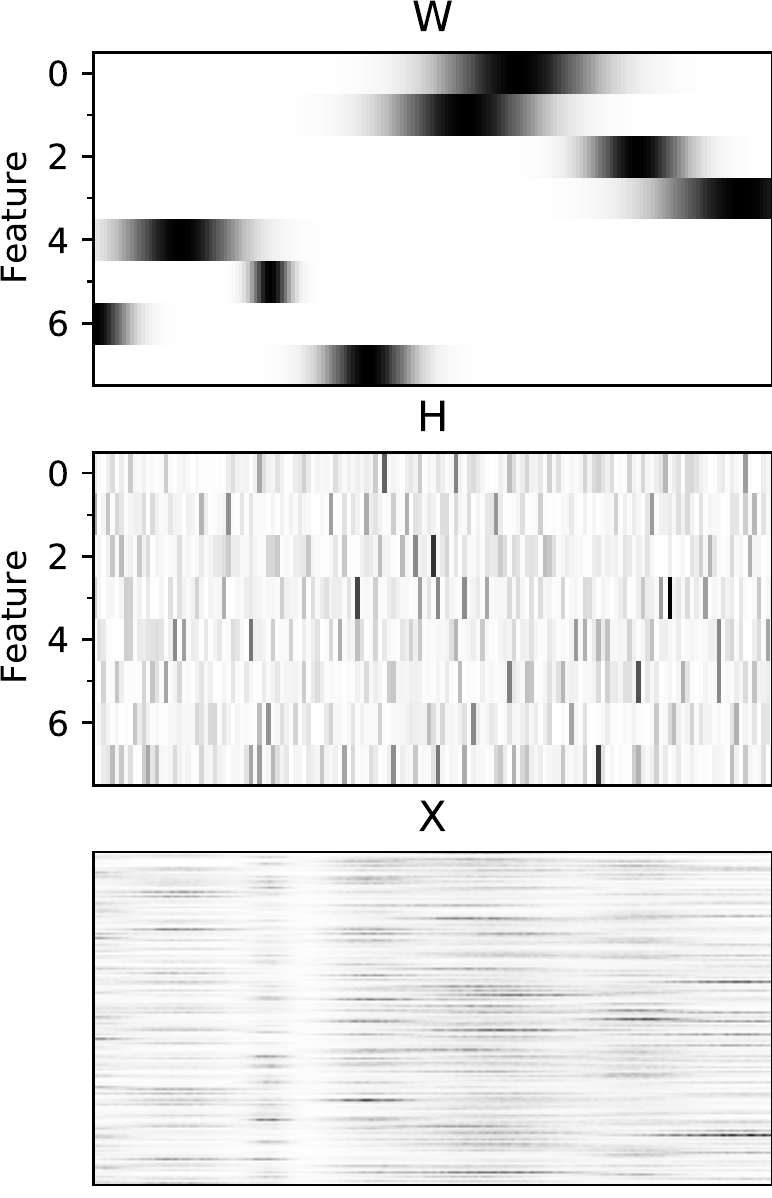}
\caption{Example $\bf{W}$, $\bf{H}$ and $\bf{X}$ matrices represented graphically. $\bf{W}$ and $\bf{X}$ are represented transposed for better visualization. The first to columns of $\bf{W}$ are visually highly correlated, which is enforced by he generation algorithm.}
\label{Xex}
\end{figure}

\subsection{Model Inference}

After training the network as described in Section \ref{secClass} to the training data set constructed in Section \ref{secTrain}, the MLP network can be applied to 7 $K$ long sliding window across all possible values of latent variables. This results in a plot that can be seen in the right hand frame of Figure \ref{ExplFig}, where the correct number of latent variables is known to be 22. For low values of $K$, the MLP network mostly assigns the windows to class 6, which is correct since the known number of features is to the right of the window. In Figure \ref{ExplFig}, the class assignments for the window beginning at the $K$ value indicated by the x-axis are given by green dots. The first and second dips in minimum silhouette value do occasionally confuse the MLP network, causing it to assign a few of these early $K$ values to class 5. However, the MLP classifier makes correct classifications on all windows containing the correct number of features. This is indicated by the steadily decreasing diagonal set of green points, starting at $K=16$. Then, once the correct number of features passes to the left of the window origin, the network correctly assigns these sequences to class 0. 

A voting system is used to turn the assignments of the MLP classifier on the sliding window into a singular prediction of $K$. Central to this system is that predictions of the exact $K$ value should hold greater weight than predictions that the correct $K$ value is outside of the MLP sliding window. When the MLP classifier predicts an exact $K$ value (assigns it to class 1-5), 5 votes are added for the appropriate $K$ value (subsequently referred to as a hit). When the MLP classifier predicts a $K$ value outside of the current window, one vote is added to all appropriate $K$ values (subsequently referred to as a miss). For example, if the window was on the range $[5,12]$ and the MLP classifier assigned a class of 2, 5 votes would be given to $K=7$. If the classifier instead assigned a class of 6, 1 vote would be added to all $K=[12,K_{max}]$. The relative importance of hits vs. misses was optimized through testing with the held out test set used in MLP hyper-parameter optimization. 

\section{Model Testing}
\label{sec:test}

In order to validate our model, it is important to apply it to data separate from the training and validation data. This will include both synthetic data generated by the algorithm described in Section \ref{secTrain} as well as data sets generated by others for similar applications. 

\subsection{Synthetic Data}
\label{subsec:syn}

\begin{figure}[htbp]
\centering
\includegraphics[width=3in]{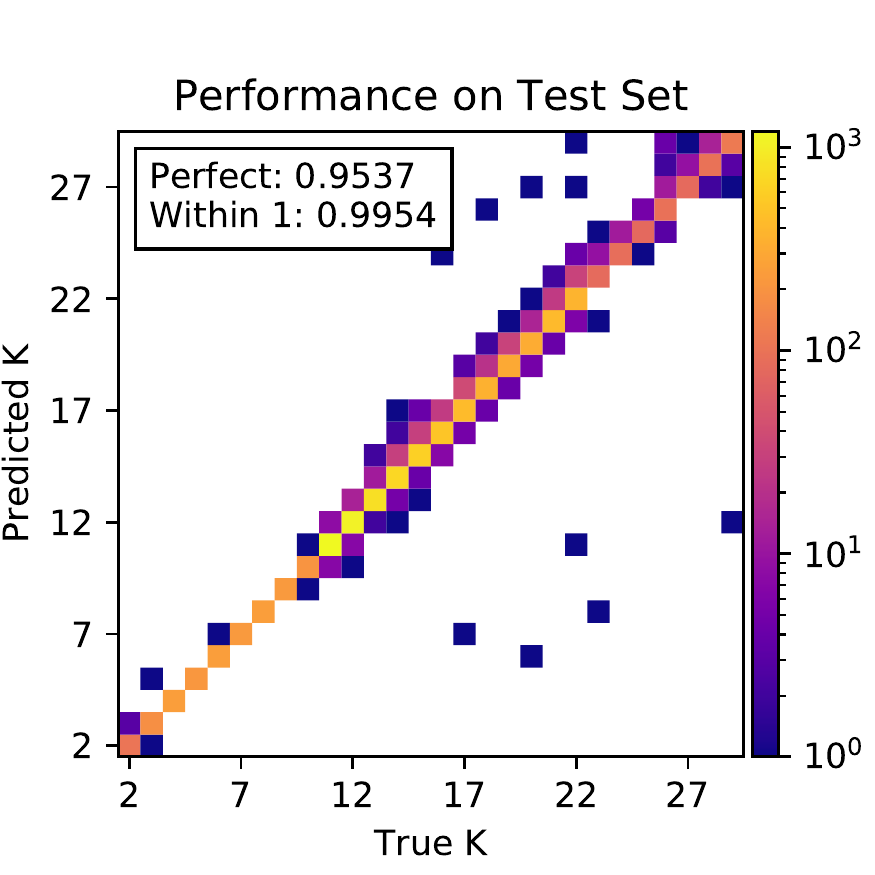}
\caption{Correlation histogram for the ML-based feature recognition algorithm versus the true number of features for the held out test set. The algorithm captures the correct number of features 95\% of the time and makes a prediction within 1 of the true number of features over 99\% of the time.}
\label{TestSetFig}
\end{figure}

The first test data set to which we apply our model is one generated in an identical fashion to the training data set, though with different random configurations. We applied NMFk algorithm to 9969 randomly generated matrices constructed in the same way as our training set. The number of features tested by NMFk was allowed to range from 1 to 36, thought the model was only allowed to predict up to a k value of 29 due to the 7 element buffer required by the model. Figure \ref{TestSetFig} shows the correlation diagram for the model predictions vs. the true K value. Remarkably, the model predicts the correct number of feature over 95\% of the time and makes a prediction within one of the correct $K$ value 99\% of the time. Additionally, the predicted number of features is within 1 of the true number of features over 99\% of the time, though this is partly due to the method of construction for the data set: only two features can be highly correlated synthetic data sets due to the generation algorithm. An illustration of how challenging assignment can be, as well as how robust the ML algorithm is, can be found in Section 1 of the Supplementary Information.

\subsection{Comparison with other methods}
\label{subsec:other}
For computational reasons, an 2679 element subset of our test synthetic data set was utilized to compare different methods of latent feature determination. This subset was limit to contain only matrices with 16 or fewer latent features. We attempted to predict the correct latent dimensionality of matrices in this data set with a variety of methods including Automatic Relevance Determination (ARD), Stability-based Coefficient\cite{Brunet4164}, and AIC. A comparison of these three methods can be seen in Figure \ref{GenTestSetFig}, with a brief description of each given in Section 3 of the Supplementary Information. Clearly, the ML-Classifier based algorithm has a significantly higher success rate at determining the correct number of features in the data set than the other methods. In this smaller data set the accuracy of the  ML-based feature recognizer increased to over 99\%, as seen in Figure \ref{GenTestSetFig}(a), likely due to the reduced number of possible $K$ values. Of the other methods, ARD is the most reliable with a success rate of 71\%. The Cophenetic Coefficient method fails on this data set, with a success rate of only 19\%.

\begin{figure}[htbp]
\centering
\includegraphics[width=6in]{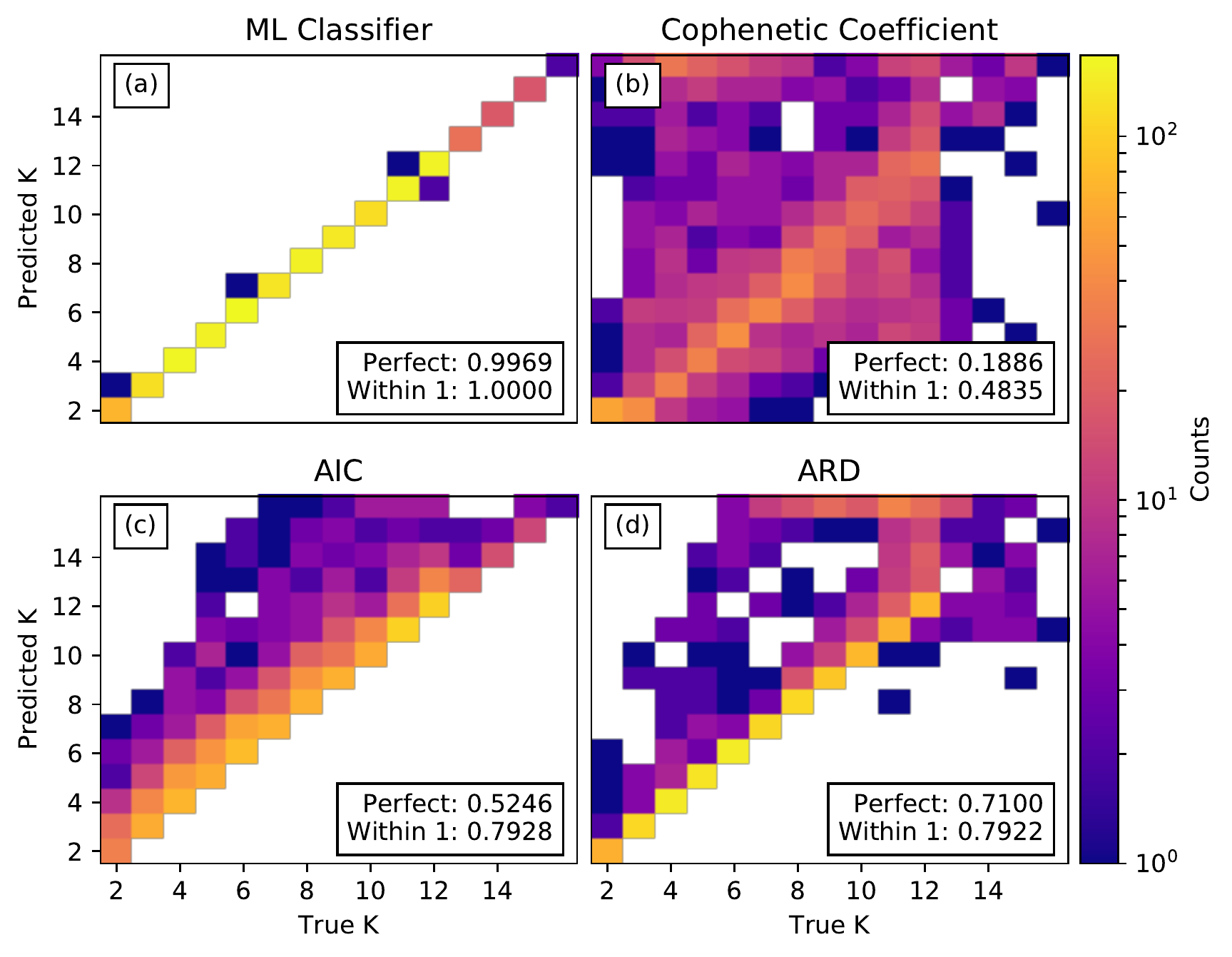}
\caption{Correlation histogram for the NMFk/MLP method, as well as ARD, AIC, and Stability-based method. Of the non ML-based methods, ARD performed the best.}
\label{GenTestSetFig}
\end{figure}

\subsection{MIT Face Data Set}
\label{subsec:MIT}

\begin{figure}[htbp]
\centering
\includegraphics[width=6in]{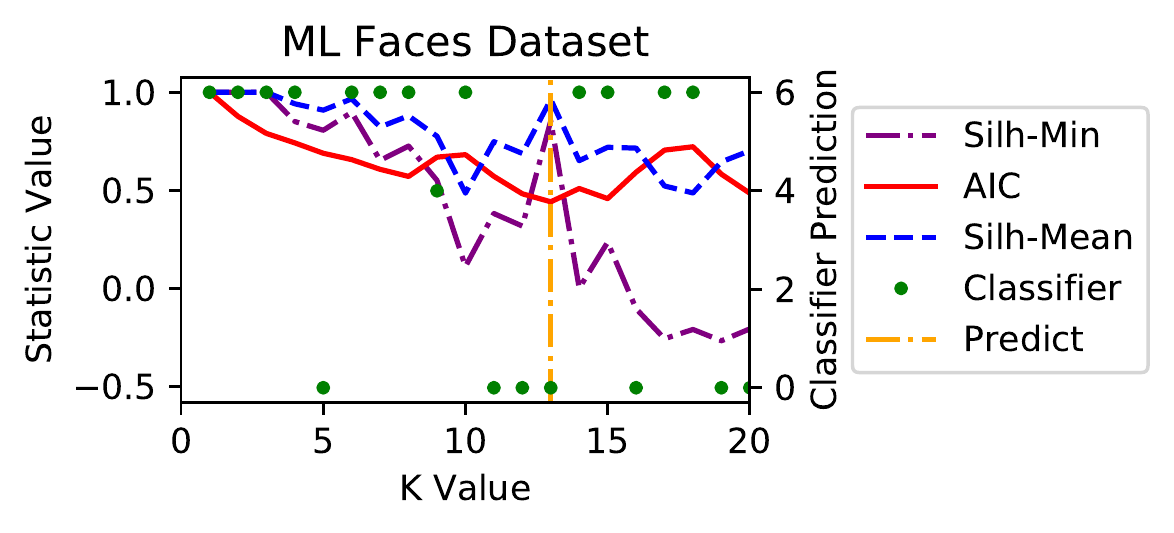}
\caption{Application of the ML feature recognition algorithm to the MIT faces data set. A reasonable $K$ value of 13 is determined by the algorithm. }
\label{FacesTestSetFig}
\end{figure}

For testing purposes, it is important to apply our method to more than just synthetic data similar to the training data set. To this end, our ML feature recognition method was applied to the face data set introduced by Lee and Seung \cite{lee1999learning}. The training subset is comprised of 2429 grey scale images of faces, each consisting of 19X19 pixels. These images were aligned by hand so that features appear at the same location in each figure. The analysis performed by our ML feature recognition algorithm is shown in Figure \ref{FacesTestSetFig}. This is a case where the ML feature classifier struggled, never only indicating the correct number of features once. When the window was positioned with the origin a $K=9$, it recognized that 13 was the correct number of features and labeled it correctly. Due to the relative importance of hits vs. misses, the algorithm settled on an answer of 13 for the number of stable features in the face data set. Importantly, this is in reasonable agreement with the ARD algorithm, which found 12 features in the face data set as shown in Ref.\cite{tan2012automatic}.

\subsection{Swimmer Data Set}
\label{subsec:swim}

Here we show the results on the Swimmer data set introduce in \cite{NIPS2003_2463}. This data set is well known to have 16 distinct features (4 positions for each of 4 limbs on a stick figure of a person). Many methods have demonstrated the ability to correctly discern the correct number of features, including ARD \cite{tan2012automatic}. With the exception of the classification of the first 5, all other windows inside of this data set were classified correctly. This includes all 7 windows containing the correct number of features, resulting in a clear diagonal line strongly indicating that 16 is the correct number of features. The initial miss-classification of the early windows was likely due to the rapid drop off of the silhouette statistic. This is another strong indicator of the practical applicability of our ML feature determination algorithm. 

\begin{figure}[htbp]
\centering
\includegraphics[width=6in]{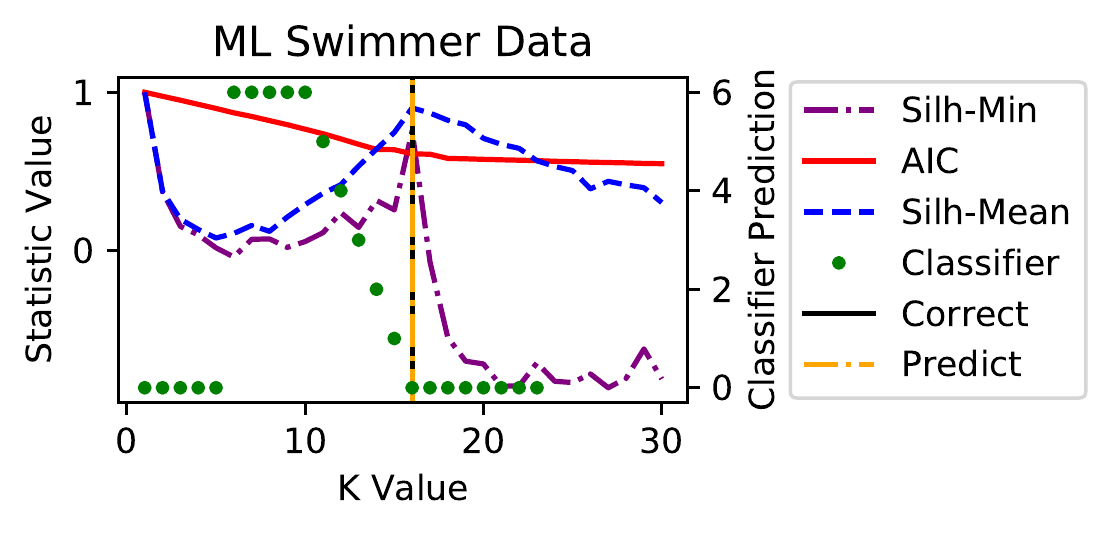}
\caption{Application of the ML feature recognition algorithm to the swimmer data set. The correct number of features is 16, which the algorithm confidently predicts.}
\label{SimmerTestSetFig}
\end{figure}

\subsection{MLIST Data Set}
\label{subsec:mnist}

As a final example of applying this algorithm to an external dataset, we chose the MNIST dataset of handwritten images \cite{dataMNIST}. This dataset consists of 70,000 hand written Arabic numerals broken into a 60,000 element training set and a 10,000 element testing set. The MNIST dataset was originally constructed for the training and testing of machines learning algorithms trained to classify hand written digits. In the present application we are interested in determining if our ML algorithm could determine a reasonable number of latent features in this image set. 

Intuitively, one would image the correct choice for the number of latent features in a database of Arabic Numerals to be 10, with each numeral getting its own feature. However, from previous literature applying NMF techniques to the MNIST dataset, a $K$ value of 10 never gives optimal results. One example of this is in the the work by Phon-Amnuaisuk who applied a dictionary based NMF technique to classify the hand written digits \cite{phonMNIST}. In this work, an NMF based classifier with $K$ values of 10, 30, 50, 70 and 90 were attempted. There is a significant increase in accuracy of the NMF classifier when moving from a $K$ value of 10 to 30, but further increases in model size have a negligible effect of performance. This indicates the the correct number of clusters in this dataset may be between 10 and 30. In work by Shan and co-workers, NMF is utilized as a dimension reduction algorithm combined with affinity propagation to determine the optimal reduced dimensionality of the dataset for an extreme learning machine based classifier. In this work, $K$ values for the data reduction NMF algorithm range from 10 to 120, with an optimal $K$ being reported as 46. This is significantly larger than both the

\begin{figure}[htbp]
\centering
\includegraphics[width=6in]{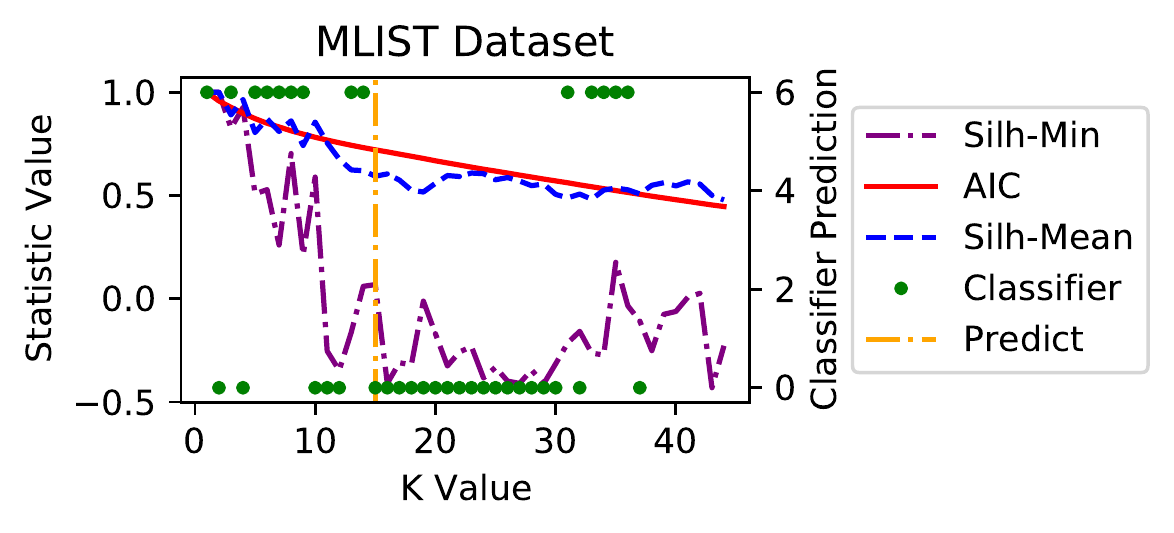}
\caption{Application to the MNIST handwriting dataset. The Silhouette statistic indicates that the correct number of features is 10, however the AIC values is steadily decreasing accross the range }
\label{mnistTestFig}
\end{figure}

When our algorithm is applied to the training fraction of this dataset, an optimal $K$ value of 15 is tentatively determined as shown in Fig. \ref{mnistTestFig}. While this is significantly lower than the value determined by Shan and co-workers, it is in good alignment with the results obtained by Phon-Amnuaisuk. Another note is that, according to the minimum silhouette statistic, there is good evidence that 10 is the correct $K$ value as that is the last large value before the minimum silhouette drops off. However the ML algorithm is likely thrown off by the constantly decreasing AIC value, preventing it from ever hitting on a precise number of features. This likely points to a need to improve the training set to account for cases where AIC becomes an unreliable indicator of correct $K$ value. 

\section{Conclusions}
\label{sec:conc}

This paper has demonstrated how fusing NMF with a MLP classifier neural network can produce a model capable of making meaningful predictions for the number of latent variables in a data set. While this problem is impossible to solve for all sets of data, this method appears to be a relatively robust method, capable of predicting the correct number of features in a variety of important data sets in the NMF community. While this is not a replacement for human data analysis, it may have a place in remote sensing operation where latent feature identification would need to be done without human intervention. Another potential application would be when there were too many data sets to be analysed individually. 

This MLP classifier can also be used to search for the correct $K$ faster, particularly when $K$ is potentially large. Current applications involve setting a lower and upper bound for $K$ and testing every value in-between. With the MLP classifier, NMFk can be run on a 7 element long window starting at a large $K$ value. Depending on whether the MLP classifier indicated a class 0 or 6 on this window, one would know whether to look for higher or lower $K$ values. In this way, the correct latent dimensionality could be determined through a divide and conquer approach. 

While this particular MLP network is not the best possible model, we have demonstrated that this type of approach to automated latent feature determination is capable of highly accurate results. As with many other ML based methods, the most critical component to building a robust model is the training data set. As such, custom training data sets could be generated to construct application specific latent feature determination networks, assuming appropriate synthetic data generators are available. Specifically, cancer applications, where NMFk has already seen extensive success and data generators already exist, is ripe for such an approach.

Alternatively, more robust MLP classifiers can be trained by augmenting the dataset with more difficult and varied matrices. Properties to change in the training dataset generation algorithm include more noisy data, latent features with varying intensities, more tightly correlated latent features, and a variety of distance metrics in the NMF algorithm. The final goal of such a robust training set would be to produce a model that correctly identifies the latent dimensionality of a data set in the vast majority of cases where NMF is able to extract meaningful features. Whether such a robust method could be developed and if it would be superior to other methods of latent feature identification such as ARD remains to be seen. 

\section{Acknowledgements}
This research was funded by Laboratory Directed Research and Development  (20190020DR), and resources were provided by the Los Alamos National Laboratory Institutional Computing Program, supported by the U.S. Department of Energy National Nuclear Security Administration under Contract No. 89233218CNA000001.

\section{Data Availability Statement}
The data that support the findings of this study are available upon request from the authors.
%The authors would like to acknowledge the Institutional Computing program at Los Alamos National Laboratory for computing resources. Additionally, the authors would like to achnowledge the Los Alamos Laboratory Directed Research and Development program for funding.  

\section{Bibliography}
\bibliography{References.bib}

\end{document}